\documentclass[lettersize,journal]{IEEEtran}
\usepackage{amsmath,amsfonts}
\usepackage{algorithmic}
\usepackage{algorithm}
\usepackage{array}
\usepackage[caption=false,font=normalsize,labelfont=sf,textfont=sf]{subfig}
\usepackage{textcomp}
\usepackage{stfloats}
\usepackage{url}
\usepackage{verbatim}
\usepackage{graphicx}
\usepackage{cite}
\usepackage{multirow} 
\usepackage{booktabs} 
\usepackage{arydshln}
\usepackage{xcolor}
\usepackage{ulem}

\begin{document}

\title{
Dual Feature Decoupling for Fine-Grained OOD Detection
}

\author{Xiaokun Li, Yaping Huang, Qingji Guan
\thanks{Xiaokun Li, Yaping Huang and Qingji Guan are from School of Computer Science and Technology, Beijing Jiaotong University, No. 3 Shangyuan Village, Haidian District, Beijing, China (E-mail: 22110102@bjtu.edu.cn, yphuang@bjtu.edu.cn and qjguan@bjtu.edu.cn).}}

\markboth{Journal of \LaTeX\ Class Files,~Vol.~14, No.~8, August~2021}%
{Shell \MakeLowercase{\textit{et al.}}: A Sample Article Using IEEEtran.cls for IEEE Journals}


\maketitle

\begin{abstract}
Out-of-distribution detection (OOD) is an indispensable technique when applying machine learning models to real-world scenarios. Most existing OOD detection methods have been developed under the idealized assumption of large inter-class distributional differences, while largely overlooking fine-grained tasks characterized by subtle variations (e.g., medical image classification, vehicle recognition). The high visual similarity among fine-grained subcategories, together with the interference of background factors, makes OOD detection extremely challenging. To tackle this problem, we propose a novel Dual Feature Decoupling Network (DFDNet), which addresses fine-grained OOD detection from the perspective of feature disentanglement. The proposed DFDNet comprises two key components: a spatial–frequency decoupling module and a reconstruction-guided decoupling module. The spatial–frequency decoupling module is designed to preserve content features that are discriminative for classification while suppressing task-irrelevant style information. on the other hand, the reconstruction-guided decoupling module introduces a novel pixel-level adversarial reconstruction task to
further remove low-level, non-discriminative information and enhance category-specific high-level semantic representations. In particular, extensive experiments demonstrate that our method achieves competitive performance improvements on multiple datasets.

\end{abstract}


\section{Introduction}
\IEEEPARstart{B}{enefiting} from the rapid advancement of Deep Neural Networks (DNNs), vision models have been widely applied in real-world scenarios. However, a visual recognition system is commonly trained in a controlled laboratory environment~\cite{7410480}, but in a real-world application, it is typically deployed in an open environment, where the data received by the system may belong to unknown categories. Therefore, to identify these unknown classes successfully, out-of-distribution (OOD) detection, which is formulated as a task of detecting whether an input data is from the training distribution (\textit{i.e.}, in-distribution, ID) or a distribution different from the training distribution (\textit{i.e.}, out-of-distribution, OOD)  has attracted more attention.

Currently, 
most existing OOD detection methods~\cite{hendrycks2017a,lin2021mood,wei2022mitigating,lee2018simple,van2020uncertainty,sastry2020detecting} primarily focus on coarse-grained scenarios, 
and the evaluation is conducted by treating different datasets as in-distribution (ID) and out-of-distribution (OOD) (e.g., CIFAR-10~\cite{krizhevsky2009learning} as ID and LSUN~\cite{yu2015lsun} as OOD). 
Owing to the relatively large semantic shift between categories from different datasets (\textit{e.g.}, truck vs. table), most state-of-the-art approaches~\cite{7410480,liang2017enhancing,lee2018simple,xiao2020likelihood,zaeemzadeh2021out,ren2021simple, wang2022vim, chen2025leveraging} can reliably detect such coarse-grained OOD samples.

\begin{figure}[t!]
    \centering
    \includegraphics[width=0.9\linewidth]{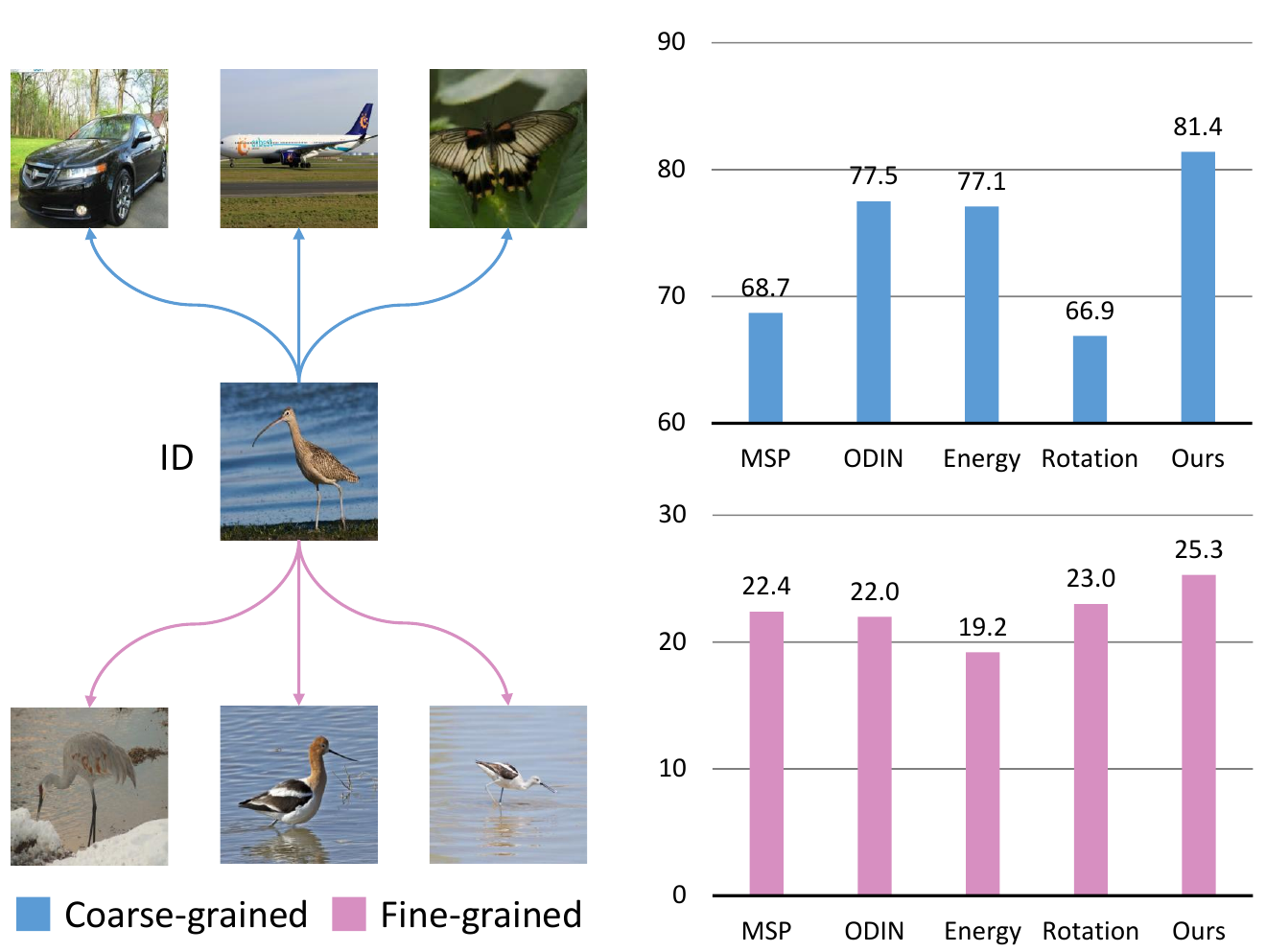}
    \caption{Comparison of OOD Detection in coarse-grained and fine-grained environments on the North American Birds dataset, respectively. Whether viewed from the perspective of human vision or deep-learning models, fine-grained OOD detection is more challenging compared to coarse-grained OOD detection.}
    \label{abstract}
\end{figure}

However, in fine-grained recognition scenarios, the input samples typically exhibit 
subtle intra-class variations, because both ID and OOD samples belong to the same coarse-grained category. As a result, they share similar global appearances and overall style features, while differing only in subtle local details. Due to the strong correlations among these fine-grained categories, classifiers trained on these data are prone to misclassification. To approximate real-world conditions, MixOE~\cite{zhang2023mixture} constructed four large-scale fine-grained datasets and demonstrated that existing OOD detection methods perform poorly in these challenging scenarios.


The reason for the unsatisfactory performance is that samples with fine-grained attributes exhibit high semantic and detail similarities, with considerable overlap in their discriminative regions, as illustrated in Fig.~\ref{abstract}. Previous methods have largely failed to extract truly discriminative features from the subtle differences between fine-grained categories. 
We believe that the key challenge in OOD detection under fine-grained settings is to disentangle task-relevant features (\textit{i.e.}, content-related and category-specific discriminative features) from task-irrelevant features (\textit{i.e.}, global style similarities such as color, texture, lighting, and other low-level details shared across regions).

To address the above issue, we propose a novel Dual Feature Decoupling Network (DFDNet) for fine-grained OOD detection. DFDNet captures distinctive representations at both the feature and pixel levels by decoupling content and style in the spatial–frequency domain and by purifying category-specific local features through reconstruction adversarial training.

Specifically, our DFDNet consists of two main modules: a spatial–frequency decoupling (SFD) module and a reconstruction-guided decoupling (RGD) module. The spatial–frequency decoupling module leverages batch normalization (BN) and instance normalization (IN) layers in a learnable manner to retain inter-class distinctiveness while suppressing style variations in spatial features. Furthermore, it incorporates the discrete Fourier transform (DFT) to eliminate additional style information and preserve the invariant characteristics of fine-grained images.
In parallel, the reconstruction-guided decoupling module decouples category-specific local features from irrelevant background by applying standard reconstruction to target regions and adversarial reconstruction to background regions, thereby reducing the influence of task-irrelevant features such as low-level details.
\begin{figure*}[ht!]
    \centering
    \includegraphics[width=0.8\linewidth]{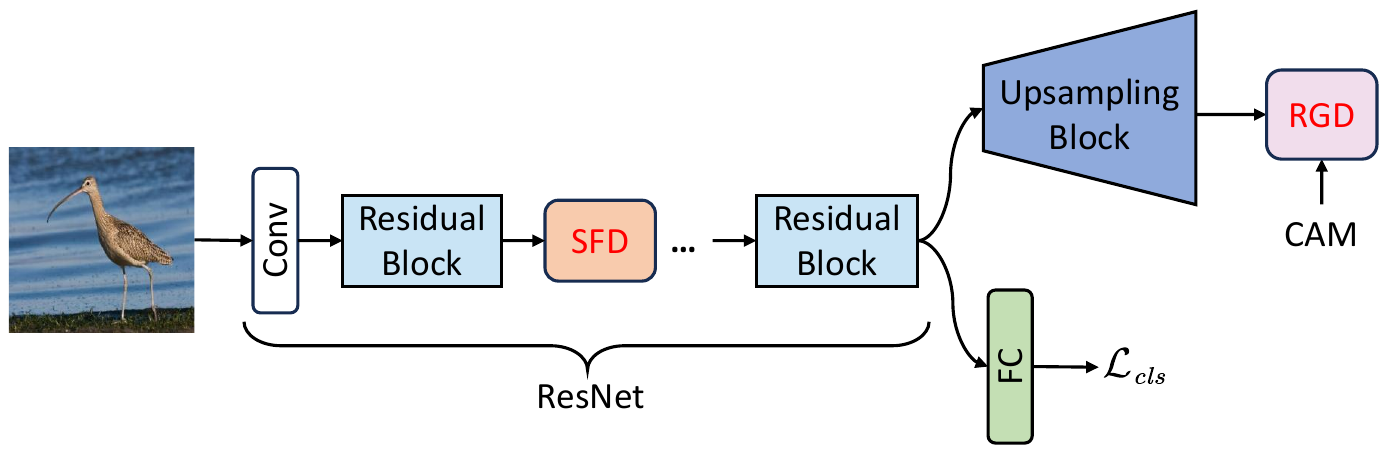}
    \caption{Overview of the DFDNet. We employ ResNet as share-weight backbone for extracting features. The spatial-frequency decoupling (SFD) modules are plugged in after residual blocks to retain content features that relevant to the classification task while removing irrelevant style features. The reconstruction-guided decoupling (RGD) module is applied in after upsampling block to eliminate low-level task-irrelevant information and purify the category-specific high-level semantic features in reconstruction process.}
    \label{overview}
\end{figure*}

In summary, the main contributions of this paper are three-fold:
\begin{itemize}
\item We address the fine-grained OOD detection from a new perspective of feature disentanglement by proposing a novel Dual Feature Decoupling Network (DFDNet), which disentangles the strong correlations between task-relevant and task-irrelevant features.

\item We design two complementary modules: (i) a spatial–frequency decoupling (SFD) module that removes style features while preserving category-discriminative features in the spatial and frequency domains, and (ii) a reconstruction-guided decoupling (RGD) module that introduces an adversarial process that integrates classification and reconstruction tasks to further purify category-specific local features.

\item We demonstrate the effectiveness of our proposed approach with several extensive experiments over four fine-grained datasets. The experimental results demonstrate that our method achieved excellent performance.

\end{itemize}

\section{Related Work}
\subsection{OOD Detection methods}
Since Maximum Softmax Probability (MSP)~\cite{hendrycks2017a} was introduced as a baseline for OOD detection, the safe deployment of models in real-world applications has received increasing research attention. Among various approaches, classification-based methods have been extensively studied, with representative techniques including post-hoc detection, confidence-enhancement strategies, and outlier-exposure methods.

Once a model is well trained, post-hoc methods can be conveniently applied. ODIN~\cite{liang2017enhancing} showed that temperature scaling combined with small input perturbations can enhance the model’s ability to distinguish OOD samples. The Energy score~\cite{liu2020energy} introduced an energy-based function as an alternative to softmax for OOD detection, where ID samples correspond to lower energy values and OOD samples to higher ones.
VIM~\cite{wang2022vim} constructed a virtual OOD logit from residual features and integrated it with the original model output to jointly determine whether a sample is out-of-distribution. DML~\cite{10204972} improved OOD detection by balancing MaxCosine and MaxNorm through the decoupling of MaxLogit. SLE~\cite{yang2025sle} exploited covariate shifts in shallow features to increase the separability of OOD samples. MIntOOD~\cite{zhang2025multimodal} hierarchically designed optimization objectives to learn robust representations that capture both coarse- and fine-grained distinctions within ID and constructed OOD data. \cite{cao2025class} employed CLIP's zero-shot capability to enhance OOD detection, effectively overcoming early stopping during prediction iterations and maintaining task recognition performance despite increasing task numbers. PRO~\cite{chen2025leveraging} leveraged the discrepancy in adversarial perturbation responses between ID and OOD samples for detection. 
Compared with post-hoc methods, confidence-enhanced approaches~\cite{devries2018learning, LU2025107427, hendrycks2019augmix,bitterwolf2020certifiably,thulasidasan2019mixup}, which are trained jointly with neural networks, aim to produce more distinguishable confidence scores for ID and OOD samples. Outlier exposure (OE) methods, such as OE~\cite{hendrycks2018deep},  G-OE~\cite{koo2024generalized}, OE-M~\cite{chen2021atom}, and MixOE~\cite{zhang2023mixture}, leveraged auxiliary OOD data as outliers to enhance the model’s ability to discriminate between ID and OOD samples.

Fine-grained OOD detection has not yet received sufficient research attention. To address this gap, MixOE~\cite{zhang2023mixture} introduced a large-scale fine-grained OOD detection dataset with hundreds of categories and proposed a training strategy that regularizes the model through mixing ID and outlier samples, thereby yielding a smoother confidence decay for OOD samples.

\subsection{Feature Decoupling}
Feature decoupling aims to decompose feature representations into independent and unrelated components. In deep learning, this strategy enables models to learn more interpretable and discriminative representations, thereby enhancing both performance and generalization. For example, DADA~\cite{peng2019domain} employed an end-to-end adversarial disentanglement framework to remove class-irrelevant and domain-specific features. MT-MIM~\cite{hou2021disentangled} disentangled identity and age components and minimizes their mutual information to mitigate the impact of age variation on recognition. DAC~\cite{lee2023decompose} combined regularization with spectral decomposition to eliminate style information while preserving content-related characteristics, thus improving domain generalization performance. CRADA~\cite{liu2024crada} addressed the UDA-OD task through cyclic reconstruction and feature decoupling in graph space, analyzing domain-specific and domain-invariant feature correlations across domains to improve feature quality. MDAL~\cite{zhang2023multi} proposed a multi-layer comparison decoupling mechanism to fuse the multi-layer features that are prone to incomplete activation and inaccurate boundaries, thereby improving object localization.

To the best of our knowledge, we are the first to introduce a feature decoupling strategy into the fine-grained OOD detection task, where feature decoupling is performed on shared discriminative regions across different subcategories. Our decoupling strategies remove style features among fine-grained categories while preserving distinctive features, thereby improving fine-grained OOD detection performance.

\section{Methodology}

We consider a $K$-class classification model trained on a fine-grained dataset ${\mathcal{D}^{in}}=\{(x_{i}^{in},y_{i}^{in})\}_{i=1}^{M}$, where $y_{i}^{in}\in \{1,\dots,K\}$ denotes the label of sample $x_{i}$ and $M$ is the total number of training samples. The OOD dataset ${\mathcal{D}_{out}}$ $\{(x_{j}^{out},y_{j}^{out})\}_{j=1}^{N}$ consists of $N$ samples, where each $x_{j}^{out}$ denotes an OOD input image and $y_{j}^{out}$ belongs to an OOD label set $\{K+1, K+2, \dots\}$. Note that the in-distribution and out-of-distribution label sets are disjoint, \textit{i.e.,} $y_{i}^{in} \cap y_{j}^{out} = \emptyset$. Our objective is to determine whether an input belongs to the in-distribution or is out-of-distribution. Fig.~\ref{overview} illustrates the overall architecture of our proposed Dual Feature Decoupling Network (DFDNet). Specifically, we insert Spatial–Frequency Decoupling (SFD) modules after the residual blocks at different stages and further incorporate a Reconstruction-guided Decoupling (RGD) module after the upsampling block. By introducing dual decoupling strategies, we can remove irrelevant features among fine-grained categories for discrimination.

\subsection{Spatial-frequency Decoupling Module}
\begin{figure}[t!]
    \centering
    \includegraphics[width=0.9\linewidth]{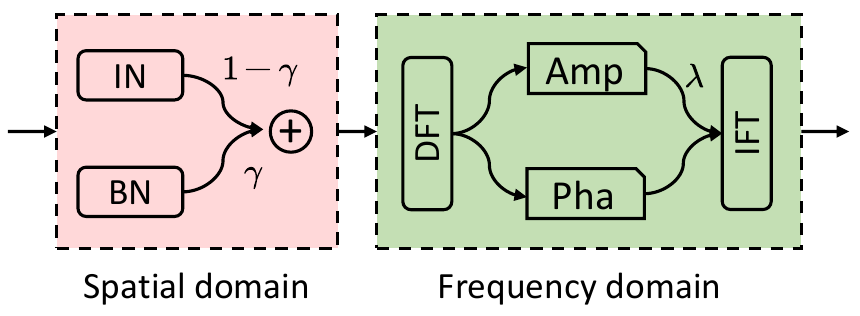}
    \caption{Illustration of the proposed Spatial-Frequency Decoupling (SFD) module. IN and BN are fused in the spatial domain with a learnable weight $\gamma$, while DFT-based decomposition regulates amplitude (style) and preserves phase (content) with a learnable parameter $\lambda$ in the frequency domain.}
    \label{SFD}
\end{figure}

For an image classification task, a classification model utilizes the extracted feature representations to determine whether the samples belong to seen categories or not. 
However, only a subset of these characteristics is visually relevant to the classification task, while characteristics related to style, such as color, texture, and lighting, are irrelevant. These style features are not directly useful for distinguishing between seen and unseen categories and often exhibit significant variations across different images.
This entanglement of relevant and irrelevant features becomes particularly problematic in fine-grained image recognition scenarios. Here, discriminative regions corresponding to different subcategories frequently overlap. These regions often contain not only task-relevant semantic features but also task-irrelevant style attributes, making separation more challenging.
To this end, we introduce the spatial–frequency decoupling module, as illustrated in Fig.~\ref{SFD}, which is designed to preserve task-relevant content features while suppressing irrelevant style features in both the spatial and frequency domains.


Batch Normalization (BN)~\cite{ioffe2015batch} and Instance Normalization (IN)~\cite{ulyanov2017improved} are widely adapted in style transfer tasks to achieve style transfer between different instance images in the spatial domain. In particular, instance normalization (IN) is commonly employed to align the statistical properties of each sample with those of the target style. Motivated by this, a natural strategy for suppressing style-related features in fine-grained OOD detection is to incorporate instance normalization (IN) into the learning process.
However, instance normalization (IN) can lead to a decrease in the inter-class variance of images, causing a reduction in the discriminative capability of the network. 
On the contrary, batch normalization (BN) helps to preserve task-relevant content features across samples, yet it remains highly sensitive to appearance-based variations such as changes in style, texture, or lighting. 
Clearly, batch normalization (BN) and instance normalization (IN) are complementary in nature. A more effective strategy is therefore to integrate the two normalization schemes. However, a simple concatenation is suboptimal for OOD detection, as fine-grained categories are characterized by both large intra-class variance and high inter-class similarity. 

In this work, to better reduce the intra-class variance and enlarge inter-class discrepancy in spatial domain, we propose a learnable fusion approach to combine the benefits of two normalization schemes:
\begin{equation}
    {f_{norm}}={\gamma f_{bn}+(1-\gamma)f_{in}},
\end{equation}
where $f_{norm}\in{\mathbb{R}^{H\times{W}\times{C}}}$ is the normalized feature after the fusion process, $f_{bn}$ and $f_{in}$ are calculated by BN and IN respectively, and $\gamma$ is a learnable weighting parameter.

In addition, despite IN helping the model remove style differences between samples and focus more on task-relevant content features, the style and content features may still be coupled in the regularized features since the content and style components are innately entangled and hard to be disentangled clearly in spatial domain, especially in the much difficult fine-grained environments. Therefore, we further employ the discrete Fourier transform (DFT) method to disentangle the features in frequency domain. Specifically, for a single channel feature map $f_{norm}$ from the spatial domain, its Fourier transformation $\mathcal{F}(x)$ is formulated as:
\begin{equation}
    {\mathcal{F}_{norm}(u, v)}=\frac{1}{HW}\sum_{w=0}^{W-1}\sum_{h=0}^{H-1}f_{norm}(w, h)\exp^{i2\pi(\frac{uw}{W}+\frac{vh}{H})},
\end{equation}
where $w$ and $h$ denote the spatial feature coordinates, $u$ and $v$ represent the frequency coordinates in the Fourier domain, and $i$ is the imaginary unit. $H$ and $W$ are the height and width of the feature map, respectively. $\mathcal{F}_{norm}(u, v)$ consists of complex values and can be represented by:
\begin{equation}
    {\mathcal{F}_{norm}(u, v)}=R(\mathcal{F}_{norm}(u, v))+iI(\mathcal{F}_{norm}(u, v)),
\end{equation}
where $R(\mathcal{F}_{norm}(u, v))$ and $I(\mathcal{F}_{norm}(u, v))$ are the real and imaginary parts of $\mathcal{F}_{norm}(u, v)$, respectively.

In the Fourier space, $\mathcal{F}_{norm}(u, v)$ consists of an amplitude component $\mathcal{A}_{norm}$ and a phase component $\mathcal{P}_{norm}$ as follows:
\begin{equation}
    {\mathcal{A}_{norm}}={\sqrt{R^2(\mathcal{F}_{norm}(u, v))+I^2(\mathcal{F}_{norm}(u, v))}},
\end{equation}

\begin{equation}
    {\mathcal{P}_{norm}}={arctan\frac{R(\mathcal{F}_{norm}(u, v))}{I(\mathcal{F}_{norm}(u, v))}}.
\end{equation}
The phase generally represents the information about the content of the image and the amplitude represents the information about the style. Although style information does not have semantic information, completely removing it will obviously affect performance. Therefore, we introduce a learnable parameter $\lambda$ to regulate the degree of style information suppression.
\begin{equation}
    {f_{out}}={\mathcal{F}^{-1}(\mathcal{P}_{norm} + \lambda \mathcal{A}_{norm})},
\end{equation}
where $\mathcal{F}^{-1}$  denotes the inverse Fourier transform, and $f_{out}$ represents the spatial feature obtained after the spatial-frequency decoupling (SFD) module, with irrelevant style information suppressed.

\subsection{Reconstruction-guided Decoupling Module}
\label{sec 3.2}
In general, we expect the extracted features to contain more discriminative information and less irrelevant information, thereby improving the performance of OOD detection. The spatial-frequency decoupling module is designed to suppress task-irrelevant style information in the feature space. However, due to the high similarity among fine-grained subcategories in most regions, certain task-irrelevant low-level details (e.g., similar backgrounds in bird images) may still be entangled with category-specific discriminative features in the pixel space. Therefore, further decoupling these low-level details and background information from discriminative features is also crucial for improving OOD detection performance.

Therefore, we propose a reconstruction-guided decoupling module that leverages adversarial learning to disentangle category-specific local features from irrelevant background. By introducing a reconstruction adversarial task and preventing the input images from being effectively reconstructed, the module drives the model to separate useful features from background-irrelevant ones.
However, performing adversarial learning by jointly reconstructing and classifying all pixels of the entire image
inevitably introduces task-irrelevant background information into the learning process. To mitigate this, we leverage the semantic localization capability of the classification task to focus on category-specific local regions from the input. These semantic regions are then reconstructed normally, while the background areas undergo adversarial reconstruction to prevent the model from overfitting to task-irrelevant information.

\begin{figure}[t!]
    \centering
    \includegraphics[width=0.9\linewidth]{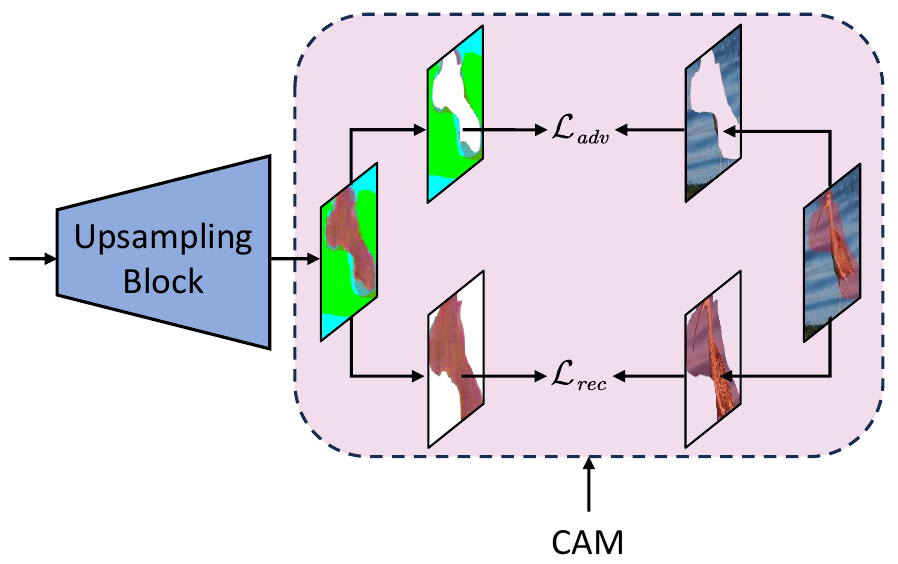}
    \caption{Structure of the reconstruction-guided decoupling (RGD) module. The category activation mapping (CAM) is applied to decouple the feature areas related to the category and pixel-wise MSE loss is applied to maintain the discriminative features, while the background area is reconstructed in an adversarial way to prevent the model from focusing on task-irrelevant features.}
    \label{RGD}
\end{figure}

Specifically, an upsampling block is appended after the feature extraction module (Fig.~\ref{overview}) to perform image reconstruction, which consists of several upsampling layers, residual blocks, and $1 \times 1$ convolutional layers, without additional specialized operations. Before backpropagation, the reconstructed image contains both foreground and background information. To decouple discriminative regions associated with the target category, we utilize a widely adopted class activation mapping (CAM)~\cite{zhou2016learning} technique to segment the image into foreground and background regions as shown in Fig.~\ref{RGD}. The class activation mapping $CAM$ is derived from the weights of the fully connected layer in the classification branch, and a threshold $\tau$ is applied to segment the target category region.

\begin{equation}
    {\mathcal{M}_{cam}}=\mathbb{I}(CAM>\tau),
\end{equation}
where $\mathbb{I}$ is an indicator function defined as $\mathbb{I}(\text{CAM}>\tau)=1$ if the value of $\text{CAM}$ exceeds $\tau$, and $0$ otherwise.

Afterwards, we first apply a conventional pixel-wise MSE loss to the target class region to improve the model's ability to learn discriminative features.

\begin{equation}
    {\mathcal{L}_{rec}}={\left \| \mathcal{M}_{cam}\cdot(x_{i} - x_{i}^{rec}) \right \| }_{2},
\end{equation}
where $x_{i}$ denotes the input image, and $x_{i}^{rec}$ denotes the image reconstructed by the upsampling block.
On the other hand, for the background regions, we adopt adversarial reconstruction to prevent the model from focusing on task-irrelevant areas.

\begin{equation}
    {\mathcal{L}_{adv}}={1-{\left \| (1-\mathcal{M}_{cam})\cdot(x_{i} - x_{i}^{rec}) \right \| }_{2}}.
\end{equation}

In summary, the reconstruction-guided decoupling (RGD) module leverages the semantic localization capability of the classification model to separate foreground and background regions. For the foreground region, the feature reconstruction is applied to enhance the model’s perception of category semantics, while for the background region, adversarial reconstruction is employed to prevent the model from focusing on task-irrelevant background information. 

\subsection{Joint Training}
As described above, our method is a multi-task learning framework comprising two branches: classification branch and reconstruction branch. The classification branch performs a supervised classification task on the extracted features, while the reconstruction branch is responsible for feature reconstruction. Specifically, the classification branch generates class activation mappings (CAMs), which are utilized to guide the reconstruction branch toward task-oriented reconstruction. In turn, the reconstruction branch with foreground–background separation enhances the classification branch by providing richer and more robust feature representations through its pixel-level reconstruction capability. This design establishes a two-way collaborative enhancement mechanism between the two branches. The overall training objective of the network is defined as follows:

\begin{equation}
    {\mathcal{L}_{total}}={\mathcal{L}_{cls} + \mathcal{L}_{rec}+ \mathcal{L}_{adv}}.
\end{equation}

\section{Experiments}

\subsection{Datasets}
In order to test the performance of DFDNet in the fine-grained OOD detection task, we follow MixOE ~\cite{zhang2023mixture} to evaluate on four public fine-grained visual classification (FGVC) datasets, including FGVC-Aircraft~\cite{maji2013fine}, Stanford Cars~\cite{krause20133d}, Butterfly~\cite{10.1145/3240508.3240523}, and North American Birds~\cite{van2015building}, for a fair comparison. In addition, WebVision 1.0~\cite{li2017webvision} is used as an auxiliary training outlier dataset. A detailed introduction to each dataset is as follows:

\textbf{FGVC-Aircraft}~\cite{maji2013fine} comprises 10,200 images across 102 categories, with each category containing 100 images. We select 90 categories in the training set as ID, and the remaining ones are fine-grained OOD categories at test time.

\textbf{Stanford Cars}~\cite{krause20133d} has a total of 196 categories and contains 16185 pictures, of which the training set has 8144 pictures and the test set has 8041 pictures. 150 categories in the training set are used as ID categories, while the other 46 categories are used as fine-grained OOD categories.

\textbf{Butterfly}~\cite{10.1145/3240508.3240523} consists of 25,279 images of butterflies, covering 200 common species, 116 genera, 23 subfamilies and 5 families. We keep 150 categories as ID and the rest 50 categories are held out from the training set and considered OOD at test time.

\textbf{North American Birds}~\cite{van2015building} involves about 48,000 annotated images of 400 bird species, including 23,929 training images and 24,633 testing images. We pick 200 categories in the training set as ID, and the rest 55 categories are used as fine-grained OOD categories at test time.

\textbf{WebVision 1.0}~\cite{li2017webvision} is an auxiliary training outlier set for MixOE training, which captures the natural images of ImageNet's 1000 categories from the Internet. In order to avoid introducing images of the same category as ID and OOD, only 1948K images are filtered out and used as the outlier set. Because the epochs of fine-tuning training is relatively small, only about 70K images are used during training.

\begin{table}[t!]
  \setlength{\tabcolsep}{2.5pt}
  \centering
  \caption{The number of images for each split of four datasets are shown below.}
  \begin{tabular}{lccccc}
    \toprule
    \multirow{2}{*}{\textbf{Dataset}} &\multirow{2}{*}{Split} &\multicolumn{3}{c}{ID} &\multirow{2}{*}{Fine-grained OOD}  \\
    \cmidrule(lr){3-5}
     & &train &validation &test & \\
    \midrule
    \multirow{3}{*}{FGVC-Aircraft} &1 &5396 &604 &3000 &333\\
     &2 &5380 &618 &3002 &331\\
     &3 &5400 &597 &3003 &330\\
    \midrule
    \multirow{3}{*}{Stanford Cars} &1 &5664 &623 &6210 &1831\\
     &2 &5657 &625 &6197 &1844\\
     &3 &5574 &633 &6131 &1910\\
    \midrule
    \multirow{3}{*}{Butterfly} &1 &7072 &798 &11513 &3458\\
     &2 &6813 &735 &11019 &3952\\
     &3 &6941 &743 &11219 &3752\\
    \midrule
    \multirow{3}{*}{North American Birds} &1 &7377 &809 &8547 &2149\\
     &2 &7278 &812 &8449 &2247\\
     &3 &7219 &798 &8446 &2250\\
    \bottomrule
  \end{tabular}
  \label{dataset}
\end{table}

\subsection{ID / OOD Setting}
ID setting refers to known categories in the dataset used for training, while OOD setting refers to categories that have not been seen during training and only appear as out-of-distribution categories during testing. We further divide these OOD categories into fine-grained OOD and coarse-grained OOD categories.

\textbf{ID setting}. In order to solve the bias problem that may occur when splitting ID and OOD samples, each dataset is divided into three equal categories of ID/OOD splits. We follow MixOE's\cite{zhang2023mixture} division of specific image numbers for training, validation, and test sets for each ID/OOD split as shown in Tabel~\ref{dataset}. 

\textbf{OOD setting}. Considering the fine-grained attributes of each dataset, a subset of categories in the training set is designated as ID, while the remaining categories are treated as fine-grained OOD. For the selection of coarse-grained categories, for example, the categories from the North American Birds dataset as ID, and those from the other three datasets as OOD.

\subsection{Evaluation Metrics}
We use three metrics for evaluation: \textit{TNR95}, \textit{AUROC} and \textit{Acc}.
\textit{TNR95} measures the proportion of actual negative cases that are correctly identified as negative, while maintaining a specificity level of 95\%.
\textit{AUROC} quantifies the overall ability of a model to distinguish between two classes (\textit{i.e.,} positive and negative) by plotting the True Positive Rate (Sensitivity) against the False Positive Rate (1 - Specificity) at various decision thresholds, and the AUROC score provides a single value that represents the area under this curve, ranging from 0 to 1.
\textit{Acc} is a common performance metric used in classification tasks to measure the proportion of correctly predicted instances out of the total instances in a dataset.

\begin{table*}[ht!]
  \setlength{\tabcolsep}{2.5pt}
  \renewcommand{\arraystretch}{1.3}
  \centering
  \caption{OOD Detection performance is reported in terms of TNR95 and AUROC, with comparisons to state-of-the-art OOD detection approaches under the fine-grained setting. Methods listed above the dotted line rely solely on in-distribution data, while those below the dotted line additionally leverage an auxiliary outlier dataset for training. It is worth noting that all results are averaged over three splits.}
  {\footnotesize
  \begin{tabular}{cccccccccc}
    \toprule
    \multicolumn{2}{c}{\multirow{2}{*}{\textbf{Method}}}  &\multicolumn{2}{c}{FGVC-Aircraft} & \multicolumn{2}{c}{Stanford Cars} & \multicolumn{2}{c}{Butterfly} & \multicolumn{2}{c}{North American Birds} \\
    \cmidrule(lr){3-4} \cmidrule(lr){5-6} \cmidrule(lr){7-8} \cmidrule(lr){9-10}  
    & &TNR95 &AUROC &TNR95 &AUROC &TNR95 &AUROC &TNR95 &AUROC \\
    \midrule
    MSP\cite{hendrycks2017a} &ICLR'2017 &71.2 / 21.4 &95.0 / 80.7 &87.4 / 56.1 &97.7 / 90.3 &88.5 / 32.8 &94.7 / 80.4  &68.7 / 22.4 &93.4 / 77.1 \\
    ODIN\cite{liang2017enhancing} &ICLR'2018 &82.4 / 20.4 &96.7 / 79.8 &98.8 / 44.5 &99.5 / 88.6 &95.5 / 33.2 &95.6 / 78.1 &77.5 / 22.0 &95.1 / 76.0 \\
    Energy\cite{liu2020energy} &NeurIPS'2020 &83.0 / 20.3 &96.8 / 79.4 &99.4 / 43.8 &99.7 / 87.5 &95.4 / 30.6 &95.5 / 77.5  &77.1 / 19.2 &95.1 / 75.0 \\
    Rotation\cite{ahmed2020detecting} &AAAI'2020 &62.0 / 21.6 &93.2 / 80.9 &89.0 / 53.9 &98.0 / 90.2 &87.6 / 32.0 &94.0 / 80.5  &66.9 / 23.0 &93.2 / 77.2 \\
    VIM\cite{wang2022vim} &CVPR'2022 &\textcolor{red}{100.0} / 7.3 &99.7 / 75.3 &99.9 / 17.4 &99.9 / 81.7  &\textcolor{red}{98.0} / 22.6 &99.0 / 78.4 &84.8 / 8.9 &97.5 / 71.3 \\
    Scale\cite{xu2023scaling} &ICLR'2024 &89.5 / 12.0 &97.8 / 79.4  &99.7 / 40.3 &99.9 / 86.9 &98.4 / 33.0 &\textcolor{red}{99.2} / 80.0 &83.6 / 15.7 &96.9 / 75.8 \\
    PRO-MSP\cite{chen2025leveraging} &CVPR'2025 &76.4 / 15.8 &95.8 / 77.1  &86.6 / 41.8 &97.4 / 87.4  &91.7 / 26.5 &98.0 / 78.8 &66.7 / 16.5 &92.7 / 74.7 \\
    \hdashline
    OE\cite{hendrycks2017a} &ICLR'2018 &98.8 / 20.1 &99.7 / 80.5 &99.9 / 52.5 &100.0 / 89.9 &93.4 / 31.0 &98.5 / 79.9 &98.0 / 21.4 &99.5 / 76.1 \\
    OE-M\cite{chen2021atom} &ECML-PKDD'2021 &99.0 / 18.3 &99.8 / 80.8 &99.9 / 51.2 &100.0 / 89.9  &92.1 / 31.8 &98.5 / 79.9  &98.4 / 20.1 &99.5 / 76.1 \\
    EnergyOE\cite{liu2020energy} &NeurIPS'2020 &99.7 / 22.4 &\textcolor{red}{99.8} / 80.8  &100.0 / 46.2 &100.0 / 89.5 &97.7 / 31.0 &98.5 / 79.9 &\textcolor{red}{99.0} / 19.1 &\textcolor{red}{99.5} / 76.1 \\
    MixOE-$line$\cite{zhang2023mixture} &WACV'2023 &91.2 / 27.5 &98.2 / \textcolor{red}{84.1}  &99.8 / 63.0  &99.6 / 92.2 &94.9 / 38.5  &94.9 / 81.4 &86.3 / 26.7 &97.1 / 79.7 \\
    MixOE-$cut$\cite{zhang2023mixture} &WACV'2023 &99.2 / 29.4 &99.7 / 82.7 &99.9 / 68.9 &99.9 / 92.9 &93.9 / 40.2 &98.7 / \textcolor{red}{82.7} &91.9 / 26.7 &98.4 / 79.6 \\
    \midrule
    Ours (MSP) &- &81.1 / 24.7 &96.4 / 82.5 &92.8 / 54.2 &98.4 / 90.1  &92.2 / 31.0 &98.3 / 80.0 &81.4 / 25.3 &96.4 / 79.3 \\
    $\Delta$ &- &+9.9 / +3.3 &+1.4 / +1.8 &+5.4 / -1.9 &+0.7 / -0.2  &+3.7 / -1.8 &+3.6 / -0.4 &+12.7 / +2.9 &+3.0 / +2.2 \\
    Ours (MixOE-$cut$) &- &99.4 / \textcolor{red}{30.6} &99.4 / 82.0  &\textcolor{red}{100.0} / \textcolor{red}{70.3} &\textcolor{red}{100.0} / \textcolor{red}{93.1} &95.9 / \textcolor{red}{41.5} &98.9 / 82.2 &95.1 / \textcolor{red}{30.8} &99.0 / \textcolor{red}{81.4} \\
    $\Delta$ &- &+0.2 / +1.2 &-0.3 / -0.7 &+0.1 / +1.4 &+0.1 / +0.2  &+2.0 / +1.3 &+0.2 / -0.5 &+3.2 / +4.1 &+0.6 / +1.8 \\
    
    \bottomrule
  \end{tabular}}
  \label{performance}
\end{table*}

\subsection{Implementation Details}

In the experiments, we train DFDNet using SGD with a batch size of 32 for 90 epochs. The initial learning rate for each component of the network is set to 0.001 and decayed using a cosine annealing schedule~\cite{loshchilov2016sgdr}. For post-training scoring methods, we utilize the weights of the MixOE baseline model~\cite{zhang2023mixture} as pre-trained weights to evaluate their performance. Additionally, we fine-tune the pre-trained DFDNet baseline model for 10 epochs using auxiliary outlier data with a batch size of 16.

\subsection{Comparisons with Other State-of-the-art Methods}


\textbf{Results on fine-grained OOD detection.} We compare the proposed method with several leading OOD detection approaches, including MSP~\cite{hendrycks2017a}, ODIN~\cite{liang2017enhancing}, Energy~\cite{liu2020energy}, Rotation~\cite{ahmed2020detecting}, VIM~\cite{wang2022vim}, Scale\cite{xu2023scaling}, PRO~\cite{chen2025leveraging}, OE~\cite{hendrycks2017a}, OE-M~\cite{chen2021atom}, EnergyOE~\cite{liu2020energy}, and MixOE~\cite{zhang2023mixture}, on four fine-grained datasets. The results are summarized in Table~\ref{performance}. Compared to post-training methods, DFDNet (MSP-based) achieves the best performance on the FGVC-Aircraft and North American Birds datasets without using any additional data. On the Stanford Cars dataset, DFDNet improves TNR95 and AUROC by 1.4\% and 0.2\%, respectively, over MixOE-$cut$, outperforming all other methods. In the North American Birds dataset, which exhibits the most fine-grained attributes, DFDNet achieves significant gains of 2.9\% in TNR95 and 2.2\% in AUROC compared to MSP. Even without incorporating auxiliary outlier data, our method surpasses all approaches except MixOE in final accuracy.

\noindent \textbf{Results on coarse-grained OOD detection.} Although our method is designed for fine-grained OOD detection, it is also suitable for coarse-grained scenarios. For validation, we perform additional experiments on four datasets under the coarse-grained setting. The results are summarized in Table~\ref{performance}. 
Based on the comparative results, our method demonstrates strong advantages in the coarse-grained OOD detection setting. Without using any auxiliary outlier data, our MSP-based approach significantly outperforms traditional post-training methods across multiple fine-grained datasets, most notably achieving a remarkable improvement of 12.7\% in TNR95 and 3.0\% in AUROC on the challenging North American Birds dataset. When enhanced with auxiliary data, our method further elevates performance, surpassing state-of-the-art techniques such as MixOE-\textit{cut} with gains of up to 4.1\% in TNR95 and 1.8\% in AUROC on the same dataset. The consistent performance improvements across diverse fine-grained domains, coupled with particularly outstanding detection capability on datasets with highly refined categories, highlight our method's robustness, generalization strength, and superior handling of fine-grained attributes in coarse-grained OOD detection scenarios.

\noindent \textbf{Results on ID classification accuracy.} Our method demonstrates competitive in-distribution (ID) classification accuracy compared to existing OOD detection approaches. As summarized in Table~\ref{accuracy}, our MSP-based variant achieves an average accuracy of 88.4\% across the four fine-grained datasets, slightly outperforming standard MSP (88.1\%) and other strong baselines such as EnergyOE (88.3\%) without using auxiliary outlier dataset. More importantly, when integrated with MixOE-\textit{cut}, our approach obtains the highest average accuracy of 89.8\%, surpassing all compared methods—including MixOE-\textit{cut}(89.2\%) and MixOE-\textit{line} (89.0\%). In particular, our method achieves the best performance in the Stanford Cars (94.0\%) and North American Birds (84.6\%) datasets, further confirming its effectiveness in maintaining high ID classification performance while improving OOD detection capability.


\begin{table}[t!]
  \setlength{\tabcolsep}{2.5pt}
  \centering
  \caption{Comparison with other OOD detection methods in terms of ID classification accuracy. Avg represents the mean comparison with other methods across three splits.}
  \begin{tabular}{lccccc}
    \toprule
    \textbf{Method} & Aircraft & Stanford Cars & Butterfly & NA Birds &Avg\\
    \midrule
    MSP\cite{hendrycks2017a} &89.6 &91.8 &89.0 &82.1 &88.1\\
    Rotation\cite{ahmed2020detecting} &88.5 &91.3 &88.7 &82.0 &87.6\\
    OE\cite{hendrycks2017a} &89.2 &91.6 &88.1 &82.4 &87.8\\
    OE-M\cite{chen2021atom} &89.3 &91.1 &88.2 &82.7 &87.8\\
    EnergyOE\cite{liu2020energy} &89.2 &91.8 &89.7 &82.3 &88.3\\
    MixOE-$line$\cite{zhang2023mixture} &90.4 &92.9 &89.3 &83.4 &89.0\\
    MixOE-$cut$\cite{zhang2023mixture} &90.1 &92.9 &\textcolor{red}{90.1} &83.5 &89.2\\
    \midrule
    Ours (MSP) &89.3 &91.8 &88.8 &83.7 &88.4\\
    Ours (MixOE-$cut$) &\textcolor{red}{90.6} &\textcolor{red}{94.0} &90.0 &\textcolor{red}{84.6} &\textcolor{red}{89.8}\\
    \bottomrule
  \end{tabular}
  \label{accuracy}
\end{table}

\begin{table}[t!]
  \setlength{\tabcolsep}{2.5pt}
  \centering
  \caption{Ablation study in terms of TNR95 and AUROC on North American Birds dataset. The numbers before and after slash respectively represent the performance of the methods on coarse-grained and fine-grained samples. SFD, DFT, and RGD refer to the Spatial-Frequency Decoupling module, Discrete Fourier Transform, and Reconstruction-guided Decoupling module, respectively.}
  \begin{tabular}{lccccccc}
    \toprule
    \multirow{2}{*}{\textbf{Model}} &\multicolumn{4}{c}{SFD}  &\multirow{2}{*}{RGD} &\multirow{2}{*}{TNR95} &\multirow{2}{*}{AUROC} \\
    \cmidrule(lr){2-5}
      &\multicolumn{1}{c}{BN} &IN &$\gamma$BN+(1-$\gamma$)IN &DFT & \\
    \midrule
     \#1 &&&&&&68.7 / 22.4 &93.4 / 77.1\\
     \#2 &\checkmark &&&&&70.2 / 23.2 &94.4 / 79.1\\
     \#3 &\checkmark &\checkmark &&&&73.3 / 25.5 &94.9 / 80.0 \\
     \#4 & & &\checkmark &&&76.8 / 24.8 &95.5 / 79.9 \\
     \#5 & & &\checkmark &\checkmark &&78.3 / 24.2 &95.8 / 80.0 \\
     \#6 && &\checkmark &\checkmark &\checkmark &81.4 / 25.3 &96.4 / 79.3 \\
    \bottomrule
  \end{tabular}
  \label{ablation}
\end{table}

\subsection{Ablation Studies}
To better explore the impact of each module in DFDNet, we conduct detailed ablation studies as shown in Table~\ref{ablation}.

\textbf{The effect of each modules.} The standard MSP is used as the baseline model (model \#1). The Spatial-Frequency Decoupling (SFD) module decouples features in both the spatial and frequency domains. To validate its design, we further conduct a detailed component-wise analysis. First of all, the detection performance of both coarse-grained and fine-grained are improved after using BN (model \#2), showing that discriminative features are preserved more completely. After adding IN (model \#3), we observe enhanced effectiveness in removing style-related features for fine-grained OOD detection.
Especially introducing learnable parameter fusion (model \#4) can effectively preserve content-relevant discriminative features while removing task-irrelevant features. In order to further remove the style information, we add the Fourier decoupling (model \#5), which further improves the coarse-grained detection performance about 1.5\%. To further enhance the discriminative capability of the model, we introduce a Reconstruction-guided Decoupling (RGD) module, which eliminates low-level task-irrelevant details and background information in an adversarial learning framework (Model \#6). Experimental results demonstrate that this module contributes to an average improvement of 1.1\% in TNR95 for fine-grained OOD detection.

\begin{figure}[t!]
    \centering
    \includegraphics[width=1\linewidth]{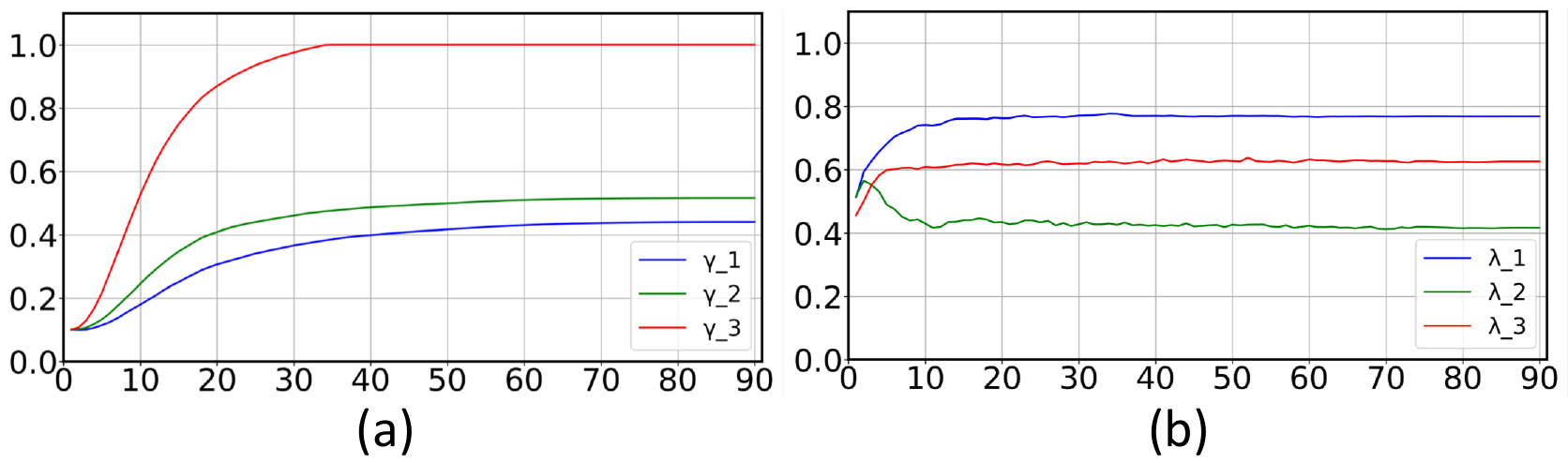}
    \caption{Experiments on the split 0 of North American Birds dataset. (a) represents the change process of the value of $\gamma$ in each spatial-frequency decoupling module during training; (b) shows that the change process of the value of $\lambda$.}
    \label{gamma_lambda}
\end{figure}

\textbf{The effect of $\gamma$ and $\lambda$.} To illustrate the influence of style information on network training, we visualize the variation of parameters $\gamma$ and $\lambda$ at different stages of the spatial-frequency decoupling module. As shown in Fig.~\ref{gamma_lambda} (a), deeper network layers correspond to progressively reduced style information within the features. Furthermore, Fig.~\ref{gamma_lambda} (b) indicates that although style information is task-irrelevant, it still constitutes an essential component of the feature representation.

\begin{figure}[t!]
    \centering
    \includegraphics[width=1\linewidth]{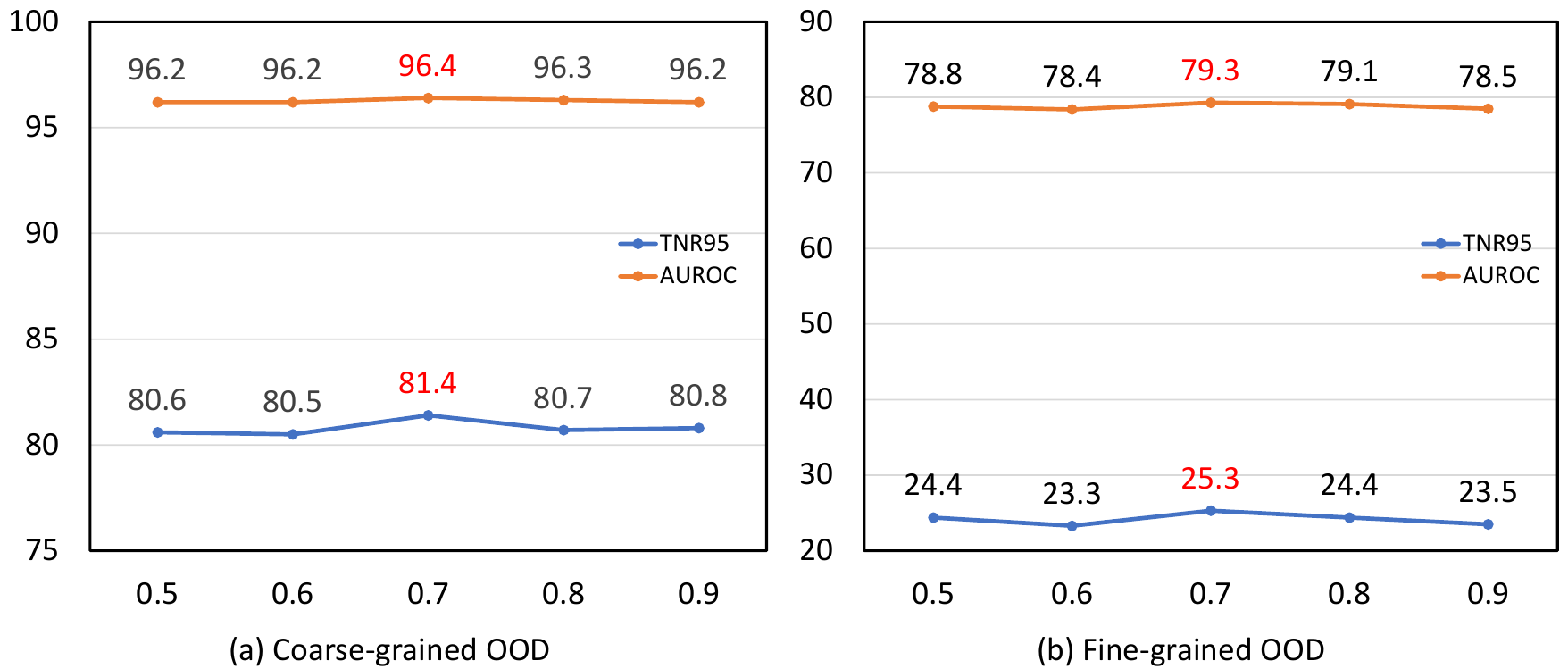}
    \caption{The influence of varying thresholds $\tau$ in class activation mappings on model performance.}
    \label{CAM}
\end{figure}

\textbf{The effect of $\tau$.}
We employ class activation mapping (CAM) in the reconstruction-guided decoupling module to decouple the target category regions from the background regions. Specifically, standard reconstruction is applied to the target category regions to enhance the model’s learning of discriminative features, while adversarial reconstruction is applied to the background regions to prevent the model from focusing on irrelevant background information. Fig.~\ref{CAM} illustrates the impact of different thresholds $\tau$ of class activation maps on model performance. Ultimately, we select 0.7 as the threshold for decoupling the foreground and background regions.

\section{Conclusion}
In this paper, we propose a novel framework, termed the Dual Feature Decoupling Network (DFDNet), for out-of-distribution (OOD) detection in fine-grained environments. Specifically, DFDNet consists of a spatial-frequency decoupling module and a reconstruction decoupling module. The spatial-frequency decoupling module preserves inter-class discriminative features while removing irrelevant style information through spatial-domain regularization, and further eliminates task-irrelevant style features via Fourier-based decoupling in the frequency domain. To separate low-level details and background information from discriminative features, the reconstruction decoupling module integrates classification and reconstruction tasks in an adversarial manner. Extensive experiments validate the effectiveness of our approach in fine-grained OOD detection tasks.

\bibliographystyle{IEEEtran}
\bibliography{reference}
\end{document}